\documentclass[runningheads]{llncs}
\usepackage{xcolor} % for defining textcolor at \todo command

\usepackage[T1]{fontenc}
\usepackage{graphicx}
\usepackage{hyperref}

\usepackage{color}

\usepackage{cleveref}  % for \cref to smartly and consistently reference tables and figures
\usepackage{amsfonts}  % e.g. for mathbb

\usepackage{array}
\newcolumntype{L}[1]{>{\raggedright\let\newline\\\arraybackslash\hspace{0pt}}m{#1}}
\newcolumntype{C}[1]{>{\centering\let\newline\\\arraybackslash\hspace{0pt}}m{#1}}
\newcolumntype{R}[1]{>{\raggedleft\let\newline\\\arraybackslash\hspace{0pt}}m{#1}}

\usepackage{subcaption}
\usepackage{booktabs}

\begin{document}
\title{Mind the Gap Between Synthetic and Real: Utilizing Transfer Learning to Probe the Boundaries of Stable Diffusion Generated Data}
\titlerunning{Mind the Gap Between Synthetic and Real}
\author{Leonhard Hennicke\inst{1}\orcidID{0009-0009-9928-4579}\and %\\
Christian Medeiros Adriano\inst{1}\orcidID{0000-0003-2588-9937} \and 
Holger Giese \inst{1}\orcidID{0000-0002-4723-730X} \and \\
Jan Mathias Koehler$^{*,}$\inst{2}\orcidID{0009-0009-5944-4849},   
Lukas Schott$^{*,}$\inst{2}
}
% %
\authorrunning{L. Hennicke et al.} % First names are abbreviated in the running head. If there are more than two authors, 'et al.' is used.
\institute{Hasso Plattner Institute, University of Potsdam, Germany
\email{firstname.lastname@hpi.de}\\
\url{https://hpi.de/} \and
Center for AI, Robert Bosch, Germany\\
\email{firstname.lastname@de.bosch.com}\\
\url{https://www.bosch-ai.com/}
}
\maketitle              % typeset the header of the contribution
\begin{abstract}
Generative foundation models like Stable Diffusion comprise a diverse spectrum of knowledge in computer vision with the potential for transfer learning, e.g., via generating data to train student models for downstream tasks. This could circumvent the necessity of collecting labeled real-world data, thereby presenting a form of data-free knowledge distillation. However, the resultant student models show a significant drop in accuracy compared to models trained on real data.
We investigate possible causes for this drop and focus on the role of the different layers of the student model. By training these layers using either real or synthetic data, we reveal that the drop mainly stems from the model's final layers.  
Further, we briefly investigate other factors, such as differences in data-normalization between synthetic and real, the impact of data augmentations, texture vs.\ shape learning, and assuming oracle prompts. While we find that some of those factors can have an impact, they are not sufficient to close the gap towards real data. 
Building upon our insights that mainly later layers are responsible for the drop, we investigate the data-efficiency of fine-tuning a synthetically trained model with real data applied to only those last layers. Our results suggest an improved trade-off between the amount of real training data used and the model's accuracy. 
Our findings contribute to the  understanding of the gap between synthetic and real data and indicate solutions to mitigate the scarcity of labeled real data.
% 246 words :) yay!!! thanks :)
%, and support the application of synthetic data in mitigating the lack of large amounts of labeled real training data.

\end{abstract}
\section{Introduction}

Deep learning has revolutionized computer vision tasks, but training neural networks in various settings incorporates challenges, like the need for large amounts of labeled training data~\cite{bhardwaj2019edgeal} and the computational resources required for deployment~\cite{ravi2016deeplchallengeswearable,zhang2019empiricaldeeplchallenges,miotto2018deeplhealthcarechallenges}. While there are approaches mitigating these challenges; they often lack specificity, e.g., when using general publicly available datasets such as ImageNet~\cite{deng2009imagenet}; or diversity, e.g., when generating data with \textit{GANs}~\cite{goodfellow2020gans}. Foundation models~\cite{bommasani2021opportunitiesfoundation} trained on large-scale diverse internet data offer a potential solution by reducing the need for task-specific training data. However, these models are still large and require expensive hardware for efficient inference. Knowledge distillation~\cite{hinton2015distilling} is an approach to transfer the knowledge from foundation models to smaller student models. From the variety of different approaches to knowledge distillation~\cite{gou2021distillingsurvey}, we focus on the concept of data-free knowledge distillation~\cite{lopes2017datafree}.

More concretely, we leverage a generative foundation model to create task-specific training images for our student model. 
We base our paper on the pipeline of Fake It Till You Make It (FITYMI)~\cite{sariyildiz2023fake} to train models on image data synthesized by StableDiffusion~\cite{rombach2022high}. 
Surprisingly, while synthetic images are humanly almost indistinguishable from real ones, there is a significant gap in performance (accuracy) when training neural networks only on synthetic data. We aim to gain deeper insights into the approximately $23$ percentage point (pp) gap in accuracy between training fully on synthetic data~\cite{sariyildiz2023fake} vs.\ real data for ImageNet-100. 

Our contributions are threefold: (1) we find that mostly the final synthetically trained layers are responsible for the drop in accuracy from synthetic to real, e.g., within a ResNet-50 the training of all but the last two layers can be done with only synthetic data, resulting in a minor accuracy drop; 
%The bottleneck is the last layer before the classification layer (\todo{correct?} where the substantial drop is taking place. 
(2) we show that one could pre-train all but the last two layers merely with synthetic data and fine-tune the remaining layers with a fraction of real data, e.g., using 1/8 of real data on a pre-trained model drops performance by 7pp compared to 20pp when omitting pre-training on synthesized data; 
(3) we show that other factors like data-augmentations, oracle prompts, differences in data normalization, and texture vs.\ shape, while potentially helpful in reducing the gap, still do not appear to be sufficient to close the gap towards real data.

\section{Preliminaries}
\textbf{Definition 1 - Foundation Models} are deep neural networks (\textit{DNN}) models with very large numbers of trainable parameters, which, thus, require orders of magnitude more data than traditional \textit{DNN} models. Besides their prediction power, foundation models~\cite{bommasani2021opportunitiesfoundation} are more adaptable to downstream tasks.

\noindent
\textbf{Definition 2 - Stable Diffusion} a.k.a.\ latent diffusion model (\textit{LDM}) is derived from a type of generative model~\cite{rombach2022high} that employs equally weighted denoising autoencoders to predict the denoised version of input data, with an objective function minimizing the difference between the added and the predicted noise.

\begin{equation}
\label{def:stable-diffusion}
    \mathcal{L}_{LDM} = \mathbb{E}_{\mathcal{E}(x),y,\epsilon\sim\mathcal{N}(0,1),t}\left[||\epsilon-\epsilon_\theta(z_t,t,\tau_\theta(y))||_2^2\right]
\end{equation}
where $\tau_\theta$ is the specialized encoder for the additional value guiding the generation process, which is trained jointly with $\epsilon_\theta$ and $y$ is the additional value. 
By shifting computations to a perceptually analogous but lower-dimensional domain through an auto-encoder, Stable Diffusion reduces the computational complexity because it can focus on the semantically significant aspects of data.

\noindent
\textbf{Definition 3 - Knowledge Distillation} consists of transferring knowledge from a larger teacher model to a smaller student model while maintaining performance and reducing computational costs~\cite{hinton2015distilling}. Initially, the teacher model is trained on a large dataset to achieve high accuracy. Then, the student model learns to mimic the teacher's behavior using both the original dataset and the teacher's soft targets, which contain probabilistic distribution insights. A Kullback-Leibler divergence loss function guides the student model to align its output distribution with that of the teacher model.
% equation and line below can be removed to save space
% \begin{equation}
% \label{eq:kl-loss}
%     KL(\hat{y}, y) = y \cdot \log\frac{y}{\hat{y}}
% \end{equation}
% where $\hat{y}$ and $y$ are the predicted and true labels respectively.

\noindent
\textbf{Definition 4 - Data-Free Knowledge Distillation} precludes the access to the original training data for transferring knowledge from a teacher model to a student~\cite{lopes2017datafree}. Instead, one leverages either the architecture or the learned parameters of the teacher model to distill knowledge to the student. This process often involves methods such as feature matching, activation mimicking, or utilizing generative models to generate synthetic data for distillation (our approach).

\section{Related Work}

\subsection{Data-Free Knowledge Distillation}
% The gap 
Current works exhibit a significant gap on ImageNet-1k~\cite{deng2009imagenet} top-1 accuracy if only trained on synthetic data compared to real data. For instance, StableRep~\cite{tian2024stablerep} has $34.9\%$ zero-shot accuracy compared to $73.7\%$ linear probing with a ViT-B/16~\cite{dosovitskiy2020image}. 
As a point of reference, we note that CLIP, which has been trained on real data~\cite{radford2021learning}, has $68.6\%$ zero-shot accuracy on a ViT-B/16 and $59.6\%$ on a ResNet50~\cite{he2016deep}. 
In contrast, SynthCLIP~\cite{hammoud2024synthclip} only provides a zero-shot accuracy of $30.5\%$ with a ViT-B/16 backbone. Lastly, also~\cite{sariyildiz2023fake} only achieves $42.9\%$ with a ResNet50. From our perspective, this gap in accuracy is quite puzzling, as for humans, synthetic images from models like StableDiffusion look almost indistinguishable from real ones. %Next, we provide more details on the methods. 

% Nearest Neighbor
We base our paper on the pipeline of Fake It Till You Make It (FITYMI)~\cite{sariyildiz2023fake}. Analogously, to their approach training models on image data generated by StableDiffusion~\cite{rombach2022high}. We adapt the FITYMI setup for our experiments to make our results directly comparable. We re-compute our replication of the paper's best-performing model for ImageNet-100 in all our result tables. Note that due to random seeds, we observe minor fluctuations ($< 1\%$  compared to the results in Table 1 in~\cite{sariyildiz2023fake} with a prompt scheme consisting of class and definition $p_c =$"$c, d_c$", and guidance scale $=2$). Furthermore, we focus on using the ImageNet-100 dataset instead of ImageNet-1K due to resource constraints, similar to most experiments in FITYMI. 
Most importantly, this reduced dataset still exhibits a fairly low accuracy of just $28.4\%$ if it is trained in a naive training fashion, e.g., without augmentations. 
% ImageNet100 of $64.8\%$ when trained on synthetic data (compared to $86.6\%$ when trained on real data), even if already many tricks like augmentations are applied. A vanilla training without any tricks, leads to a baseline performance of just $28.4\%$.
To investigate parts of this gap, the FITYMI paper already covers some ablations on ImageNet-100 like variations of prompts, augmentations, and the guidance scale that each significantly impact the performance. Especially, introducing augmentations improves the baseline performance by $14.8$ pp. Similarly, tuning the guidance scale improves the performance by $20$ pp. Note that both factors have been tuned in isolation and the benefits might be less than their sum if combined. While the aforementioned tricks increase the performance, a large portion of the gap remains, e.g., a vanilla training on the real ImageNet-100 data leads to $87.4\%$ top-1 accuracy compared to the best data-free variant from~\cite{sariyildiz2023fake} achieving $64.8\%$. Thus, we extend the ablations by testing out further factors and investigating where in the network discrepancies between real and synthetic data originate.  

Similarly, to FITYMI, StableRep~\cite{tian2024stablerep} presents a pipeline learning representations solely based on synthetic images. They use captions from the datasets CC3M~\cite{sharma_conceptual_2018}, CC12M~\cite{changpinyo_conceptual_2021} or RedCaps~\cite{desai_redcaps_2021} respectively to generate image data with Stable Diffusion v1.5 and then train the model on the generated data in an unsupervised manner to learn the representations used for linear probing and zero-shot classification. 
% Their approach achieves high top-1 accuracy on linear probing, however, it has to be noted, that these datasets include more samples than the synthetic ImageNet-1K version that is used to learn the representations in FITYMI. 
While the synthetic ImageNet-1K only includes less than 1.3 million samples, StableRep uses 10 million and 11.6 mio.\ samples from CC12M and RedCaps respectively. For CC3M only 2.7 mio.\ samples are used. Also, StableRep trains a ViT-B/16 ($\sim 86$ mio.\ trainable params.) for 35 epochs, whereas we train a ResNet50 ($\sim 25$ mio.\ trainable params.) for 100 epochs.  
% The authors of StableRep also provide results for a ViT-B/16 trained for 105 epochs, which achieves top-1 accuracies of 75.7\% and 76.7\% for CC12M and RedCap respectively.

Lastly, Azizi et al.~\cite{azizi_synthetic_2023} use Imagen~\cite{saharia2022imagen} instead of StableDiffusion to generate synthetic training data for image classification. They achieve a top-1 accuracy of 69.24\% on ImageNet using a ResNet50 trained for 90 epochs; however, they achieve this by fine-tuning Imagen on ImageNet data, resulting in their approach not being completely data-free.
\subsection{Transfer Learning}
\paragraph{Transfer Learning - \textit{TL}.} Reusing knowledge from pre-trained models via TL requires that data from pre-trained tasks (source) and new ones (target) have some overlap. Unfortunately, this assumption is unrealistic in domains subject to data sparsity~\cite{chang2019disjoint}. To bridge this gap between source-target labels, one can combine similarity-guided training and adaptive adversarial noise selection~\cite{wang2023efficient}. Another approach is to apply self-supervision methods, for instance, Yamaguchi et al.~\cite{yamaguchi2022transfer} investigated a two-stage process that (1) synthesizes target samples by conditioning on source generative model and, then, (2) labels these new samples via self-supervised learning methods. Albeit \textit{TL} has shown to outperform knowledge distillation~\cite{yamaguchi2022transfer}, one still has to cope with the threat of "negative transfer", i.e., learning wrong concepts or forgetting correct ones~\cite{zoph2020rethinking}.
\paragraph{Catastrophic forgetting - \textit{CF}.} Re-training models needs to cope with the fundamental phenomenon of catastrophic forgetting~\cite{mccloskey1989catastrophic}, i.e., neural networks when trained on different tasks tend to "forget" (lose accuracy) in previously learned tasks. This phenomenon is not limited to sequential learning setups (e.g., continual learning, reinforcement learning, domain adaptation). Fine-tuning of foundation and generative models are also affected~\cite{wang2023comprehensive}, for instance, in generative adversarial networks - \textit{GANs} trained continuously~\cite{wang2024comprehensive}. Some of the mitigation to \textit{CF} involve momentum and gradient penalty~\cite{thanh2020catastrophic} or changes in the discriminator component of the \textit{GAN}~\cite{liang2018generative}. Meanwhile, in \textit{GANs} continuously trained in the context of data-free knowledge distillation (\textit{DFKD})~\cite{chen2019data}, mitigation of \textit{CF} by preserving memory buffers for past samples~\cite{binici2022preventing}, remembering previous distributions for generative replay~\cite{binici2022robust}, and keeping an exponential moving average to minimize the impact of abrupt shifts in the distribution~\cite{do2022momentum}. 
Our approach is complementary to these approaches, as we combine an architecture-based approach (fixing parameters, i.e., layers) with synthetic or real data for pre-training and then fine-tuning. Catastrophic forgetting could happen in various steps of the pipeline, for instance, the synthetic images cause the student model to forget features that were successfully learned by the teacher model. While freezing layers might mitigate this loss, it might still happen in the retrained layers, because one cannot guarantee that all fine-tuned knowledge is new knowledge.
\section{Experiments}
\subsection{Experimental Setup}
The premise of our experiments is strongly inspired by the findings of the paper Fake It Till You Make It (FITYMI)~\cite{sariyildiz2023fake}. To make our results comparable with the ones from FITYMI, all of our experiments were run using the same setup that was used by the authors, i.e., training a randomly initialized ResNet-50~\cite{he2016deep} for 100 epochs using SGD~\cite{ruder2016overview} with a momentum of 0.9 and DINO augmentations with one global and eight local crops, unless stated otherwise. 
We linearly increased the learning rate up to 0.1 for the first 10 epochs and then decayed it on a cosine schedule. In addition, temperature scaling with L2-normalized weights was used on the final classification layer. Training a model using this setup takes over 56 single GPU hours, but it can be parallelized. Our implementation is based on the implementation of~\cite{sariyildiz2022no} - which is publicly available - adapted for our classification task.
The dataset we use is ImageNet-100~\cite{tian2020contrastive}, a subset of ImageNet-1K~\cite{deng2009imagenet}. This is the same subset of classes that is used in FITYMI. We use either the original data from ImageNet-100 or a synthetic version of this dataset with 1300 images generated per class.
The synthetic images were generated using Stable Diffusion 1.4 unless stated otherwise. This was done using 50 diffusion steps and a guidance scale~\cite{ho2022stablediffguidance} of 2.0. The images were generated sized 512 x 384 and using the WordNet~\cite{miller1995wordnet} based definition prompts from FITYMI (i.e., "<class name>, <WordNet definition of class name>"). Generating a full ImageNet-100 dataset with this setup takes over 73 single-GPU hours. %, i.e., 3.65 hours using 20 V100 GPUs.
For our experiments, we report the mean and standard deviation on real ImageNet-100 validation data of the last 5 epochs for top-1 and top-5 accuracy.
\begin{figure}[btp]
    \centering
    \begin{subfigure}{.49\textwidth}
        \includegraphics[width=0.8\textwidth]{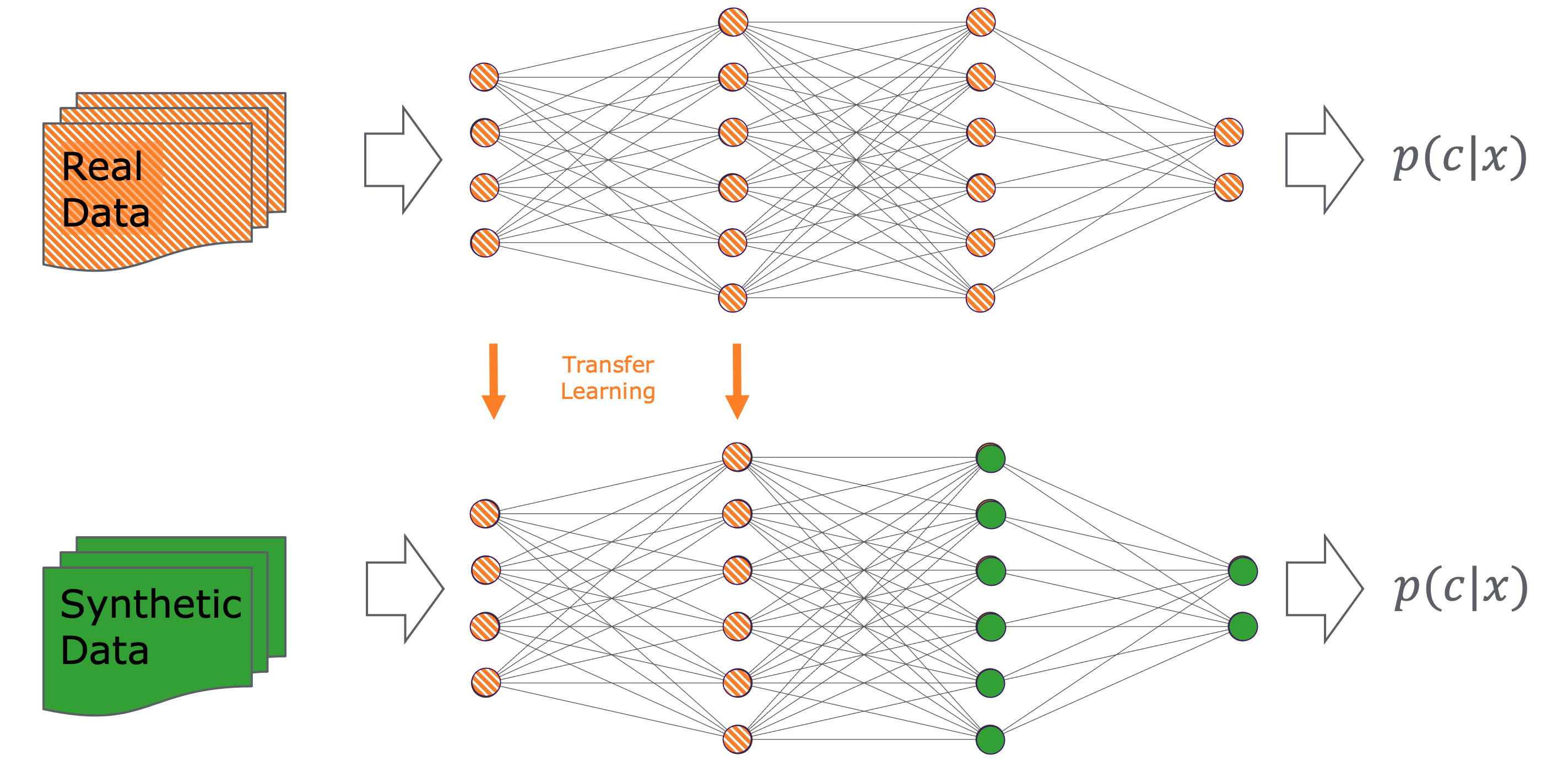}
        \caption{Real Data Transfer Learning.}
        \label{fig:real_data_transfer_illustration}
    \end{subfigure}
    \begin{subfigure}{.49\textwidth}
        \includegraphics[width=0.8\textwidth]{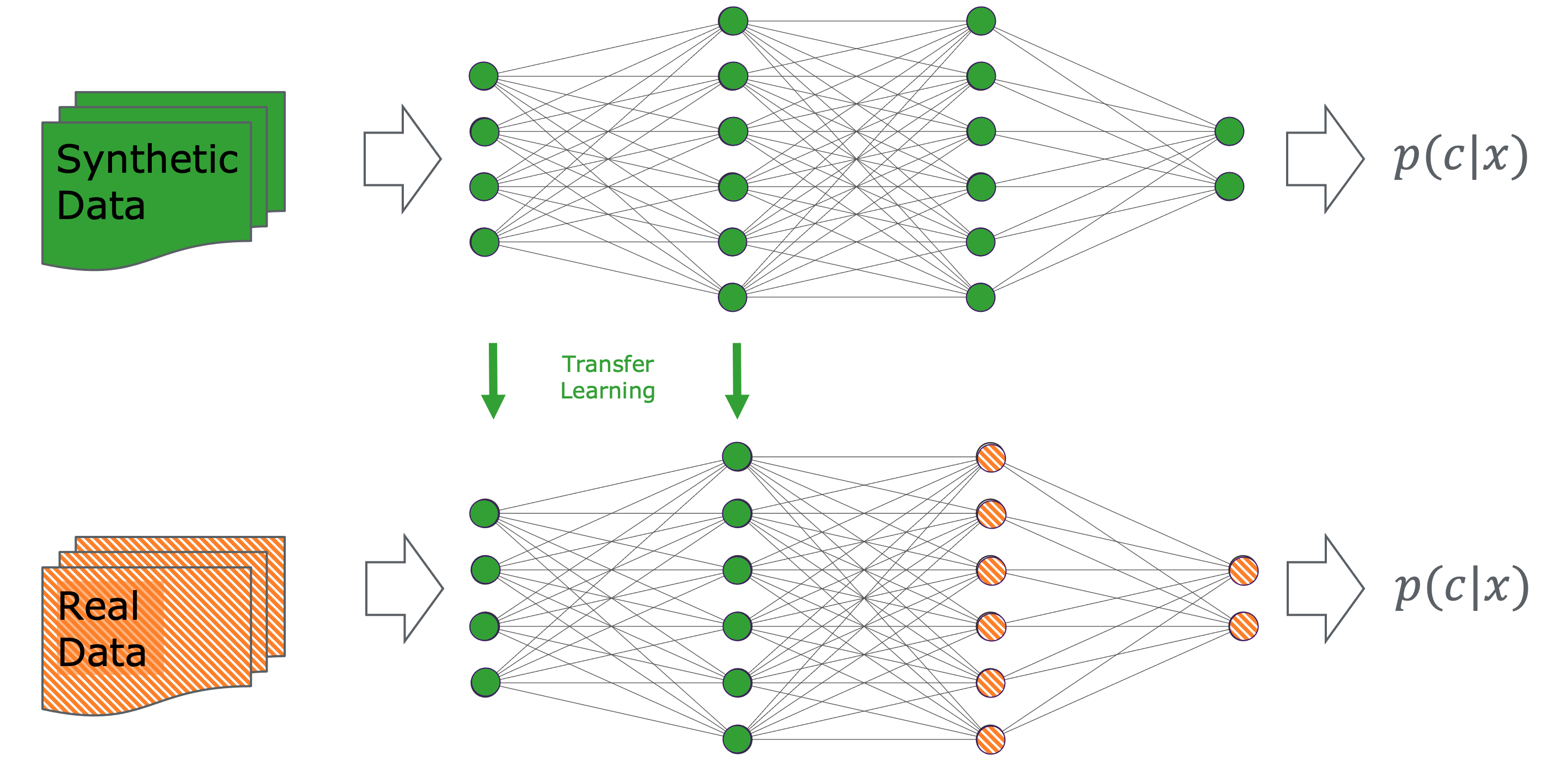}
        \caption{Synthetic Data Transfer Learning.}
        \label{fig:synthetic_data_transfer_illustration}
    \end{subfigure}
    \caption{This figure illustrates our transfer learning setup for N = 2. N indicates the number of consecutive layers that are transferred from a model that was trained on the respective first dataset, starting from the first layer.}
\end{figure}
\subsection{Layer Importance Experiments}
It is a widely accepted intuition that \textit{CNNs} (such as the ResNet-50 we are using) learn representations of more abstract features in the later layers of the model while the earlier layers learn representations of features at a lower abstraction layer~\cite{lecun2015deep}. Based on this intuition, we devised a series of experiments using transfer learning to narrow down which features are missing from the synthetic data. In these experiments, we pre-train a \textit{CNN} on either synthetic or real data and then freeze the first N layers. Next, we reinitialize the remaining layers and retrain them on the respective other dataset. As we gradually increase N, we expect to see changes in the accuracy of the resulting model, that relate to the quality of features at gradually higher levels of abstraction that are present in the data on which the model has been pre-trained. When using the term layer in the context of our experiments, we refer to the bottleneck blocks in the ResNet-50 architecture. The only exceptions to that are the first layer, which is the singular convolutional block that comes before the bottleneck blocks, and the last (18th) layer, which consists of the final (linear) classification layer and the pooling layer that precedes it. Running this setup for N = 17 results in a setup equivalent to linear probing, while N = 18 would result in only evaluating a completely pre-trained model with all parameters frozen.

\subsubsection{Real Data Transfer Learning}
In our first series of experiments, the first N layers are transferred from a model pre-trained on real data with frozen parameters, while only training the remaining layers of the model on synthetic data. We evaluate on real data. We illustrate this setup in \Cref{fig:real_data_transfer_illustration} for $N = 2$.

\begin{figure}[btp]
    \centering
    \begin{subfigure}{0.49\textwidth}
        \includegraphics[width=\textwidth]{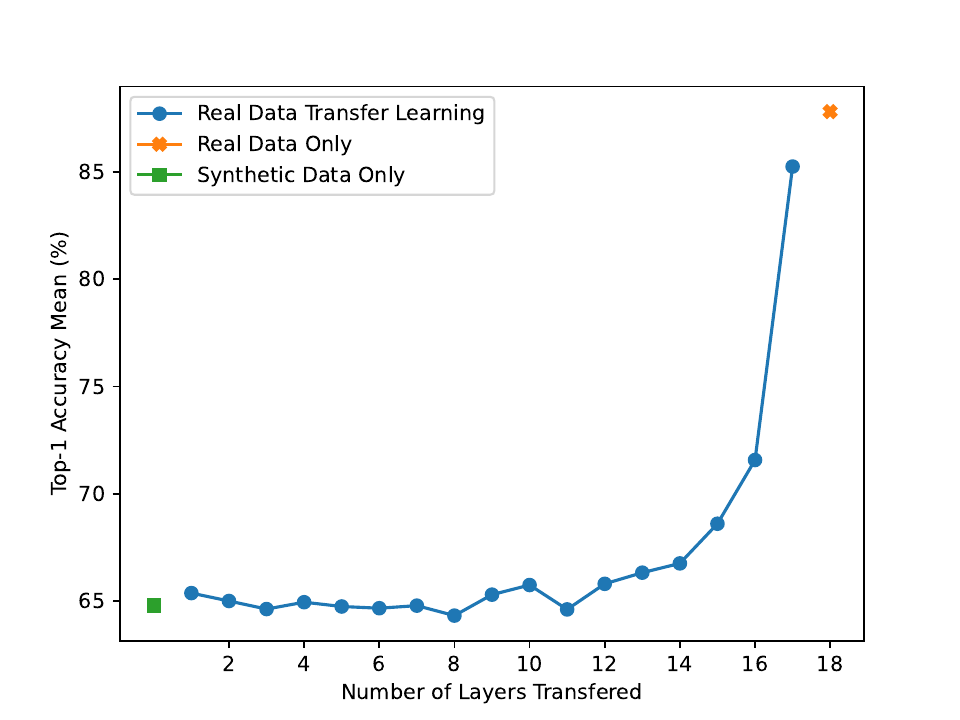}
        \caption{Real Data Transfer Learning.}
        \label{fig:abl_6_real_data_transfer_learning_plot}
    \end{subfigure}
    \begin{subfigure}{0.49\textwidth}
        \includegraphics[width=\textwidth]{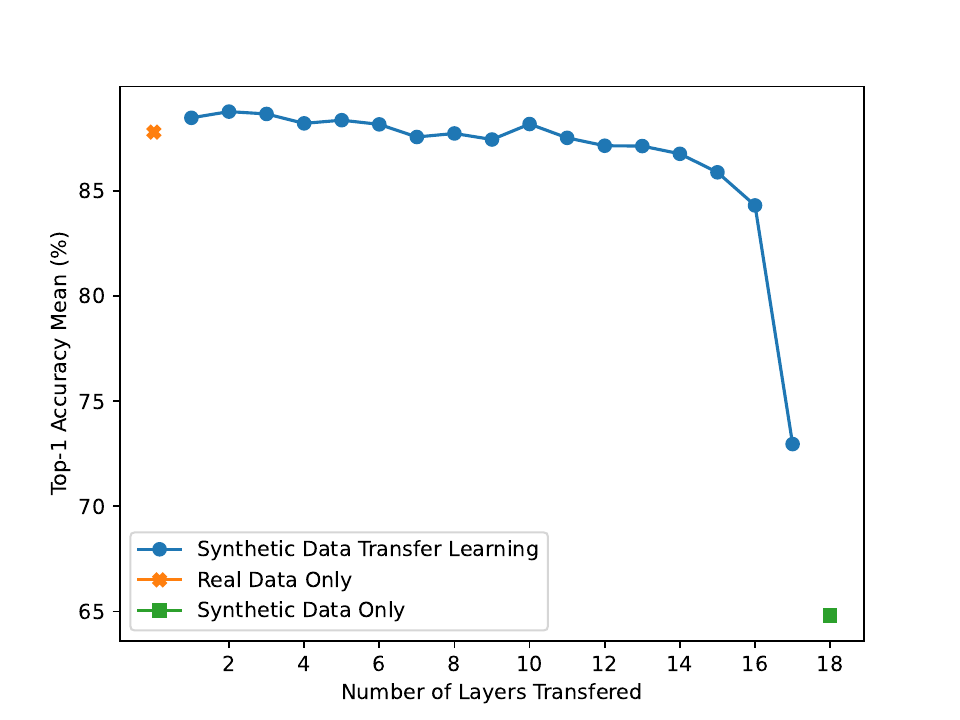}
        \caption{Synthetic Data Transfer Learning.}
        \label{fig:abl_9_synthetic_data_transfer_learning_plot}
    \end{subfigure}
    \caption{Results of the layer importance experiments using transfer learning from real and synthetic data respectively for N = 1 to N = 17. We also plot the results for our baseline models trained on synthetic and real data only, as these present the two extremes of this experiment setup and therefore provide an approximate lower and upper bound. 
    % For the full results see \Cref{tab:abl_9_synthetic_data_transfer_learning} and~\ref{tab:abl_6_real_data_transfer_learning}.
    }
\end{figure}

The results of this experiment are presented in \Cref{fig:abl_6_real_data_transfer_learning_plot}. If the representations learned by the different layers of the model were to all transfer equally well to real data, we would expect the accuracy to gradually rise as we replace more and more of the layers with the ones that were pre-trained on real data and fewer layers are trained on synthetic data. However, this is not the case, suggesting that the quality of the representations that the student model learns from the synthetic data varies depending on their level of abstraction.
As we can see in~\Cref{fig:abl_6_real_data_transfer_learning_plot}, the top-1 accuracy increases by at most 2.0pp from the baseline respectively as we add the pre-trained layers 1 to 14 and even decreases for some of the in-between layers. In contrast to that, when adding the pre-trained layers 15 to 17, we can see progressively larger increases\footnote{Here we refer to the increases relative to the model with only the previous pre-trained layers added respectively.} in top-1 accuracy of 18.4pp in total, which is visible by a steep increase for the layers 15 to 17. 
%In addition to the accuracy increases, we can also see the training loss more than double after these three pre-trained layers are added, which is suggestive of the model overfitting less.
Using the representations learned from real data by the first 14 layers makes little difference over using the ones learned from synthetic data compared to the differences using the representations learned by the later layers. 
\subsubsection{Synthetic Data Transfer Learning}
To further support our claim, we present a second series of experiments similar to the ones in the previous section. Now, the first N layers are transferred from a model pre-trained on synthetic data with frozen parameters, while the remaining layers of the model are reinitialized and trained on real data (setup illustrated in \Cref{fig:synthetic_data_transfer_illustration} for N = 2). Evaluation is always performed on real validation data.
Similarly to our first series of experiments, if the representations learned from the synthetic data by the different layers of the model were all transferred equally well to real data, we would expect the accuracy to gradually decrease
as we replace more and more of the layers with the ones that were pre-trained on
synthetic data and fewer layers trained on real data are used.
The results of this experiment are presented in \Cref{fig:abl_9_synthetic_data_transfer_learning_plot} where we plot the top-1 accuracy against the number of layers transferred. Supporting our previous findings, we can see that the accuracy varies only little when pre-training layers 1 to 16 with synthetic data. 
%in comparison to the overall gap in accuracy between the baseline models trained on real data and on synthetic data only respectively, although the entire gap has to be bridged eventually, by adding more and more pre-trained layers. 
Specifically, the top-1 accuracy only decreases by 3.5 pp, as the layers 1 to 16 are replaced by the ones pre-trained on synthetic data. In fact, the top-1 accuracy even increases up to 1.0pp as some of the in-between pre-trained layers are added. 

A larger decrease of 11.3pp in top-1 accuracy is visible as the 17th layer (the second to last layer) is replaced by one that is pre-trained on synthetic data - noticeable as a steep drop in \Cref{fig:abl_9_synthetic_data_transfer_learning_plot}. The decrease in accuracy is accompanied by an increase in training loss. This suggests that the decrease in accuracy is probably not due to the model overfitting, but rather due to the features learned from synthetic data not transferring well to the real evaluation data.

\subsection{Data-Reduction Experiments}
Based on the results of our layer importance experiments, we suspect that a model with only the last layers trained on real data (and the earlier layers were already pre-trained on synthetic data) would require less data to achieve a similar accuracy as a baseline model trained fully on real data. 
To test this hypothesis, we pre-train the first 16 layers of our model on synthetic data, i.e., $N=16$, and then reinitialize and retrain the remaining layers on a randomly drawn, reduced number of samples from the real ImageNet-100 dataset, effectively reducing the amount of real data necessary for training.
Assuming our intuition is correct, we should observe that the accuracy remains within a similar range as the baseline model accuracy, as we decrease the amount of real training data used.

\begin{figure}
    \begin{subfigure}{0.49\textwidth}
    \centering
    \includegraphics[width=\textwidth]{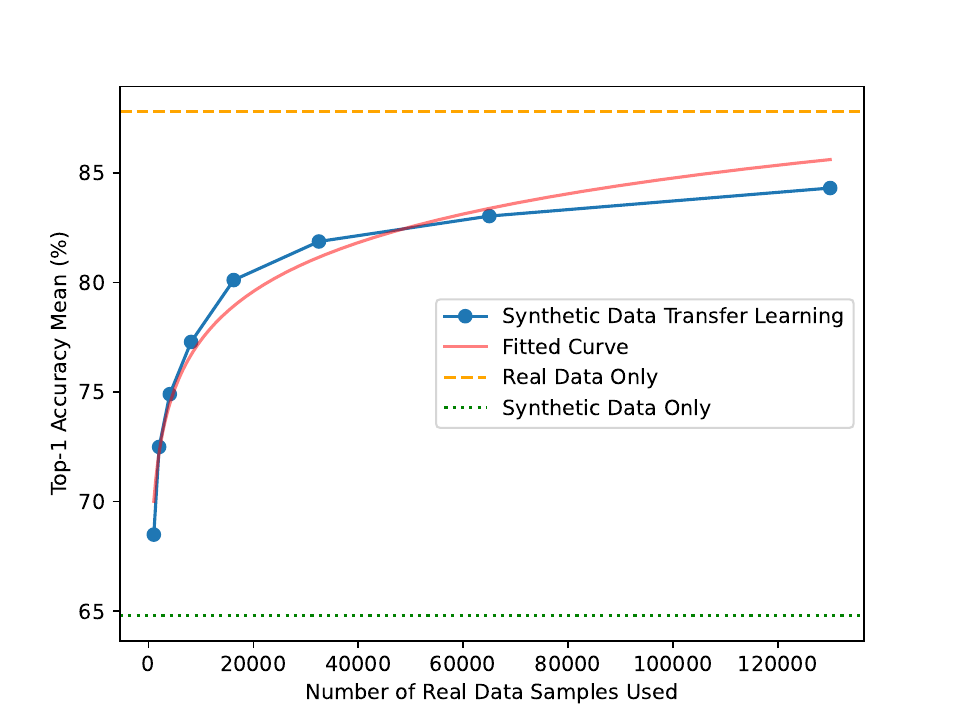}
    \caption{Linear Scale (Read Right to Left)} %$f(x) = 2.22761899x + 47.76753691$.
    \label{fig:abl_15_data_reduction_fitted_linear}
\end{subfigure}
\begin{subfigure}{0.49\textwidth}
    \centering
    \includegraphics[width=\textwidth]{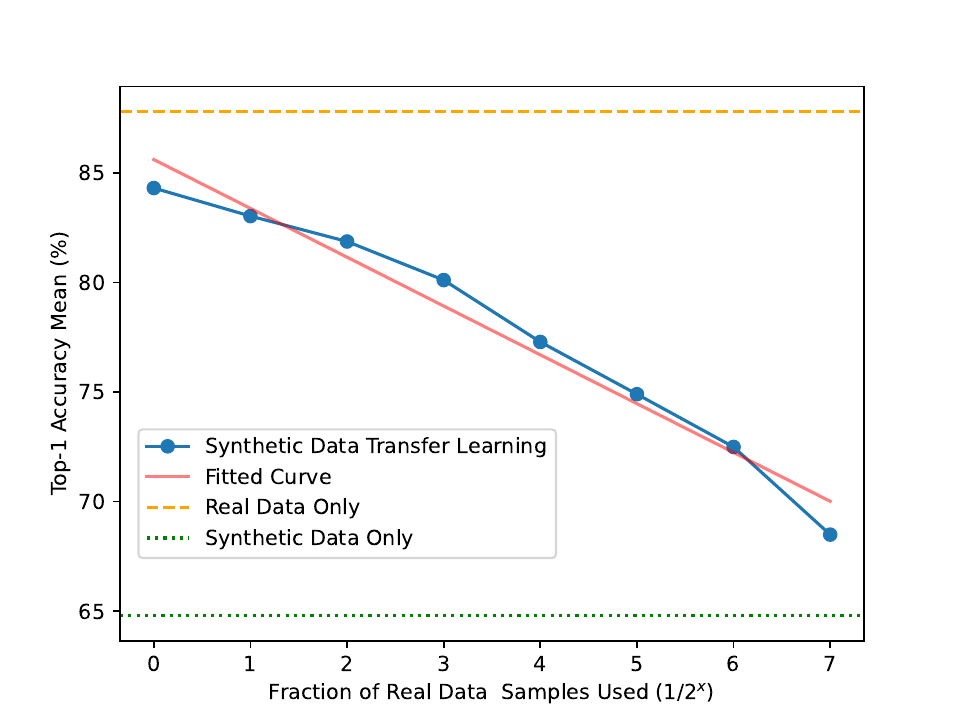}
    \caption{Logarithmic Scale (Read Left to Right)}
    \label{fig:abl_15_data_reduction_fitted_log}
\end{subfigure}
\caption{\textbf{Data Reduction Experiments} The first 16 layers are trained on synthetic data and frozen. The last two layers are fine-tuned with a different number of real data samples. 
The fitted curves are derived from the plotted top-1 accuracy on the respective scales via curve fitting, using least squares polynomial fitting on a 1st-degree polynomial ($f(x) = -2.23 log(x) + 85.61$). }
\label{fig:abl_15_data_reduction}
\end{figure}

\begin{figure}[h!]
    \centering
    \begin{subfigure}{0.49\textwidth}
        \includegraphics[width=\textwidth]{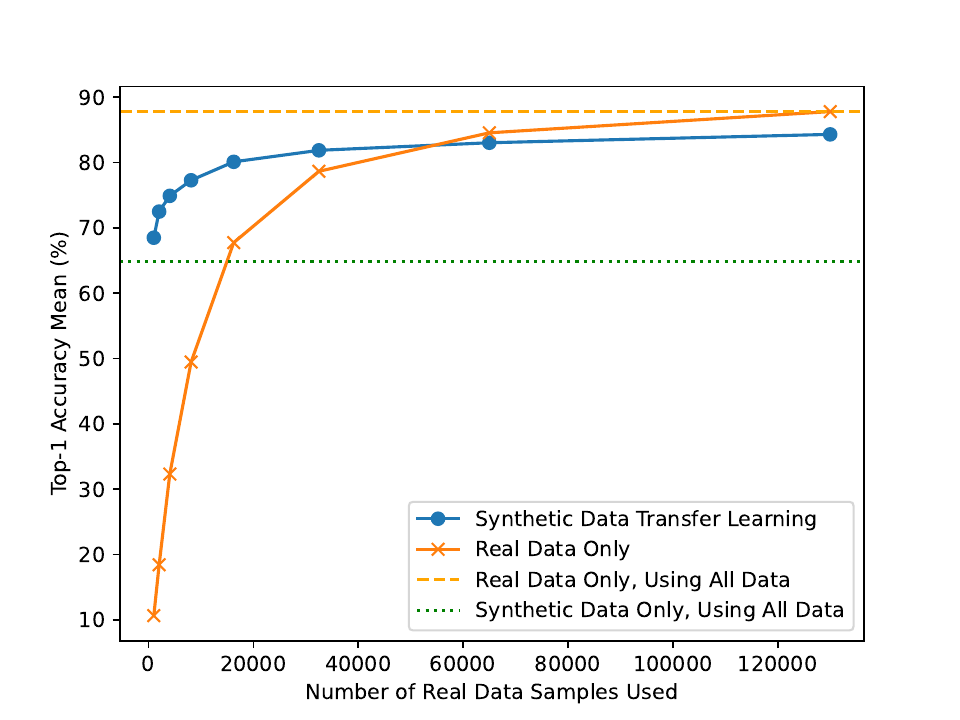}
        \caption{Linear Scale (Read Right to Left)}
        \label{fig:abl_15_data_reduction_comparison_linear}
    \end{subfigure}
    \begin{subfigure}{0.49\textwidth}
        \includegraphics[width=\textwidth]{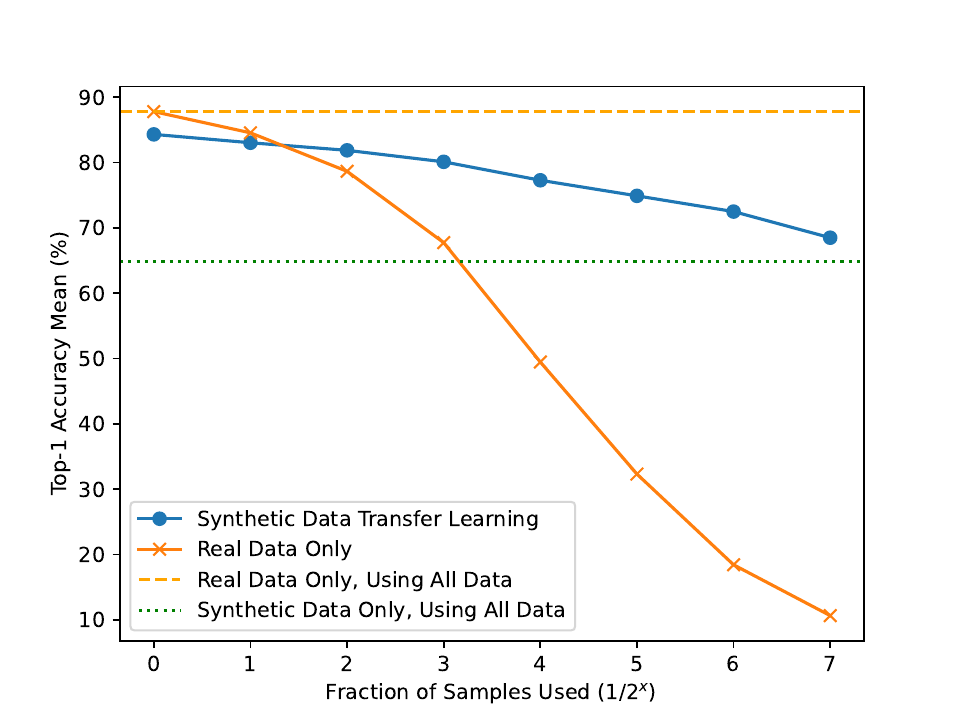}
        \caption{Logarithmic Scale (Read Left to Right)}
        \label{fig:abl_15_data_reduction_comparison_log}
    \end{subfigure}
    \caption{\textbf{Training on a reduced real dataset} Top-1 accuracy of reducing the amount of real training data with (blue, same as~\Cref{fig:abl_15_data_reduction}) and without (orange) synthetic data transfer learning. }
    \label{fig:abl_15_data_reduction_comparison}
\end{figure}

The results of these experiments are presented in \Cref{fig:abl_15_data_reduction}. We decrease the amount of real training data by halving it for each consecutive experiment. Even though we reduce the real training exponentially, the top-1 accuracy only decreases by 4.2pp after the third iteration (when using 1/8th of the entire real training data).
% In \Cref{fig:abl_15_data_reduction_fitted_log} we plot the top-1 accuracy in relation to the fraction of real training data samples used, i.e., on a logarithmic scale.
% We also plot a linear curve in the same figure, which is derived from the plotted top-1 accuracy on this scale via curve fitting, using least squares polynomial fitting on a 1st-degree polynomial. 
The fitted curve suggests that the top-1 accuracy decreases logarithmically, as the amount of real training data is reduced.
Additionally, we show in \Cref{fig:abl_15_data_reduction_comparison} a series of reductions in real training data for a randomly initialized model, the drop in accuracy is steeper than for the models with frozen layers pre-trained on synthetic data, suggesting that synthetic pre-training can be applied to mitigate the lowered accuracy caused by a lack of real data.

\subsection{Other Potential Causes of the Accuracy Gap}
% Next we present other potential causes of the accuracy gap.
%based on our findings do not seem to be contributing to the phenomenon directly.
\paragraph{\textbf{Normalization}}
%Because the synthetic data are likely to have different first and second-order statistics, normalizations used by the student model could be off when the model switches from synthetic data during training to real data during inference. 
We investigate if differences in first and second-order statistics (dataset mean and variance) of the synthetic data that is used during training, and the real data that is used during evaluation, cause the drop in performance. 
%
% \textbf{Setup: }
We tested two interventions during the training of the student model. 1) We normalize the input using the actual (channel) mean and standard deviation of the respective dataset instead of the default ImageNet values during both training and evaluation. 2) We set the batch norm layers to train mode during evaluation, causing them to keep updating their running mean and std.\ dev.\ instead of using the ones learned during training. 
%
% \textbf{Results: }
We observe a slight improvement in accuracy for these interventions on both real and synthetic data (see \Cref{tab:normalization}). However, the gap might not to be explainable by exact normalization. 
\begin{table}[btp]
\caption{\textbf{Normalization Experiments.} Exact normalization: image normalization using the actual (channel) mean / std. dev. of (R) or (S) data instead of ImageNet values during training and evaluation. Batch norm: batch normalization layers keep updating running batch mean and std.\ dev.\ during evaluation.}
\label{tab:normalization}
\begin{tabular}{ l C{2.5cm} C{2.5cm} C{3cm} }
\toprule
Experiment & Top-1 Acc. (\%) & Top-5 Acc.  (\%) & Train Loss \\
\midrule
Real Data Only (R) & 87.8 ±0.1 & 97.1 ±0.1 & 0.742 ±0.002 \\
+ Exact Input Normalization & {88.5 ±0.1} & 97.5 ±0.1 & 0.728 ±0.003 \\
+ Batch Norm in Train Mode & {88.5 ±0.1} & 97.3 ±0.1 & 0.726 ±0.003 \\
\midrule
Synth. Data Only (S) & 64.8 ±0.1 & 87.6 ±0.1 & 0.696 ±0.003 \\
+ Exact Input Normalization & {65.9 ±0.1} & 87.6 ±0.2 & 0.694 ±0.004 \\
+ Batch Norm in Train Mode & {65.2 ±0.2} & 87.3 ±0.2 & 0.700 ±0.003 \\
\bottomrule
\end{tabular}
\end{table}

\paragraph{\textbf{Data Augmentation}}
As shown in FITYMI~\cite{sariyildiz2023fake}, the type of augmentations can make a difference for models trained on synthetic data, more so than when training on real data. Based on this, we suspected that the gap could be bridged through data augmentation. 
We test several sophisticated augmentation pipelines, both for models trained on real and on synthetic data. However, the accuracy gap still remains and we could also not replicate the effect of these augmentations w.r.t.\ leading to a larger improvement when training on synthetic data (as shown in FITYMI) compared to our baseline using no data augmentations. As shown in ~\Cref{tab:augmentations}, we still observe this effect when comparing the specific augmentations used in FITYMI, i.e., the ones from the PyTorch example~\cite{pytorchExampleAugs} and the ones from DINO~\cite{caron_emerging_2021}. However, the effect is less pronounced in our results than in the ones shown in FITYMI. The difference is likely due to a lower guidance scale being used for the specific experiment in FITYMI. This suggests that the increased
improvisation of Stable Diffusion (due to the lower guidance scale) has a positive effect on the generated data that is similar to the augmentations.
\begin{table}[ht]
\caption{\textbf{Augmentation Experiments.} The models were trained using different augmentation techniques, as specified in the \textit{Experiment} column. To accurately portray the impact of the different augmentations, the baseline was trained using the FITYMI setup, as in our other experiments, but without the DINO augmentations. Hence, in this table, only the runs labeled "DINO" are equivalent to the full FITYMI setup, that we us as a baseline everywhere else.}
\label{tab:augmentations}
\begin{tabular}{ l C{2.5cm} C{2.5cm} C{3cm} }
\toprule
Experiment & Top-1 Acc. (\%) & Top-5 Acc. (\%) & Train Loss \\
\midrule
Real Data Only (R) & 70.1 ±0.1 & 88.2 ±0.2 & 0.007 ±0.001 \\
+ AutoAugment & {79.7 ±0.1} & 94.6 ±0.1 & 0.033 ±0.001 \\
+ PyTorch Example~\cite{pytorchExampleAugs} & {86.2 ±0.1} & 96.5 ±0.1 & 0.308 ±0.003\\
+ AugMix~\cite{hendrycks_augmix_2020} & {86.3 ±0.1} & 97.2 ±0.1 & 0.318 ±0.002 \\
+ PixMix~\cite{hendrycks_pixmix_2022} & {86.5 ±0.1} & 97.2 ±0.1 & 0.848 ±0.007 \\
+ DINO~\cite{caron_emerging_2021} & {87.8 ±0.1} & 97.1 ±0.1 & 0.742 ±0.002 \\
\midrule
Synth. Data Only (S) & 46.9 ±0.2 & 71.7 ±0.2 & 0.006 ±0.001 \\
+ AutoAugment & {56.2 ±0.2} & 81.4 ±0.1 & 0.026 ±0.100 \\
+ PyTorch Example~\cite{pytorchExampleAugs} & {59.5 ±0.1} & 83.9 ±0.2 & 0.261 ±0.002 \\
+ AugMix~\cite{hendrycks_augmix_2020} & {60.3 ±0.2} & 84.5 ±0.2 & 0.281 ±0.003 \\
+ PixMix~\cite{hendrycks_pixmix_2022} & {61.8 ±0.2} & 84.7 ±0.2 & 0.680 ±0.006 \\
+ DINO~\cite{caron_emerging_2021} & {64.8 ±0.1} & 87.6 ±0.1 & 0.696 ±0.003 \\
\bottomrule
\end{tabular}
\end{table}

\paragraph{\textbf{Local Textures}}
Images generated by Stable Diffusion, often have inconsistencies in finer structures (like human hands) or local textures~\cite{rombach2022stablediff}. This leads us to the question of whether local textures generated by Stable Diffusion are less useful to the student model when generalizing to real data. This would contribute to the accuracy gap, as it has been shown repeatedly, that convolutional neural networks, such as our student models are biased towards local textures~\cite{geirhos2018imagenet}.
To test this, we generated two new datasets by applying the techniques from Stylized ImageNet~\cite{geirhos2018imagenet} to our real and synthetic ImageNet-100 datasets respectively. Stylized ImageNet uses style transfer techniques in combination with paintings to remove local texture cues and in return only retain global shape information. Despite this intervention, the accuracy gap remains (see ~\Cref{tab:stylized_imagenet}). Note the lower accuracy values w.r.t. the baseline. This is  expected, as we introduce an information loss to our data by removing features from local textures.
\begin{table}[btp]
\caption{\textbf{Stylized ImageNet Experiments.} The entire training dataset was modified using the techniques from Stylized ImageNet, removing local texture cues but retaining the global shape information.}
\label{tab:stylized_imagenet}
\begin{tabular}{ l C{2.5cm} C{2.5cm} C{3cm} }
\toprule
Experiment & Top-1 Acc. (\%) & Top-5 Acc. (\%) & Train Loss \\
\midrule
Real Data Only (R) & 87.8 ±0.1 & 97.1 ±0.1 & 0.742 ±0.002 \\
+ Stylized ImageNet & {75.6 ±0.2} & 92.2 ±0.1 & 2.099 ±0.005 \\
\midrule
Synth. Data Only (S) & 64.8 ±0.1 & 87.6 ±0.1 & 0.696 ±0.003 \\
+ Stylized ImageNet & {50.5 ±0.1} & 77.0 ±0.2 & 2.053 ±0.006 \\
\bottomrule
\end{tabular}
\end{table}
\paragraph{\textbf{Prompt Optimization}}
Based on the findings in FITYMI, we asked whether the accuracy gap may be caused by unused potential in prompt engineering. To simulate optimal prompts, we applied unCLIP~\cite{ramesh2022hierarchical} to generate synthetic images with Stable Diffusion based on CLIP~\cite{radford2021clip} embeddings of real ImageNet images as oracle prompts instead of text embeddings as is in our other experiments, i.e., we generate one synthetic image for each ImageNet-100 image using the real image as an input for unCLIP. As CLIP creates a shared embedding space for text and images, this in essence simulates the perfect text description for each of the real images. 
Note that, since the Stable Diffusion (SD) model used in unCLIP is fine-tuned to also accept image embeddings, the image prompts provide more information than just perfect text prompts, as the resulting embedding vectors will have the perfect magnitude in addition to the perfect angle. In our evaluation, we switch to a synthetic dataset generated by Stable Diffusion v2.1 for a fair comparison as this is the version used by unCLIP. As seen in ~\Cref{tab:abl_4_unclip}, we do not bridge the accuracy gap with this intervention. However, compared to the lower-performing SD v2.1 baseline, we achieve a top-1 accuracy improvement of $26.3$ pp, which is larger than the original accuracy gap of $23$pp between real data and synthetic data generated by SD v1.4. Potentially, this could allow prompt engineering to bridge the accuracy gap, but, one should be careful drawing conclusions from this experiment. CLIP is trained on a contrastive loss based on cosine similarity between text and image embeddings while unCLIP is trained on CLIP image embeddings. Hence, prompts based on image embeddings could induce data leakage, as they provide more information than even a perfect text prompt would do, e.g., by having both the perfect angle and magnitude.
\begin{table}[btp]
\caption{\textbf{unCLIP Experiment.} We compare student models trained on various synthetic datasets generated by Stable Diffusion. "unCLIP" indicates that the model was trained on a dataset generated using unCLIP (based in Stable Diffusion v2.1) with the real ImageNet-100 images as prompts. SD v1.4 and SD v2.1 indicate the respective Stable Diffusion versions used to generate the dataset.}
\label{tab:abl_4_unclip}
\begin{tabular}{ l C{2.5cm} C{2.5cm} C{3cm} }
\toprule
Experiment & Top-1 Acc. (\%) & Top-5 Acc. (\%) & Train Loss \\
\midrule
unCLIP, (SD v2.1) & {66.6 ±0.3} & 90.8 ±0.1 & 0.690 ±0.004 \\
\midrule
Synth.\ Data Only (SD v2.1) & {40.3 ±0.3} & 66.3 ±0.4 & 0.511 ±0.002 \\
Synth.\ Data Only (SD v1.4) & {64.8 ±0.1} & 87.6 ±0.1 & 0.696 ±0.003 \\
\bottomrule
\end{tabular}
\end{table}

\section{Discussion}
\subsection{Implications to Transfer Learning} We showed that choosing specific layers to transfer can have a positive impact on transferring knowledge from the teacher to the student model while fine-tuning with real data. This knowledge can be used for mitigating catastrophic forgetting in situations where knowledge distillation is necessary to cope with data sparsity and to produce more compact models. Because synthetic data generation is extremely flexible, this extra control leverage can support a more fine-grained study of how certain image features in a particular class (e.g., vehicle type) or domain (e.g., weather conditions) are more prone to be forgotten given the previously learned features. Such extra flexibility could enable "negative transfer" strategies for mitigating shortcut learning~\cite{geirhos2020shortcut}, hence endowing the fine-tuning pipeline and the resulting models with robustness to non-stationary data generation processes. Ultimately, combining synthetic data with layer freezing could provide control baselines to evaluate the existing mitigation strategies against \textit{CF} in continuously trained \textit{GANs}~\cite{thanh2020catastrophic,liang2018generative}, particularly in DFKD approaches~\cite{binici2022preventing,do2022momentum}.
\subsection{Threats to Validity} Next, we discuss possible threats to the validity~\cite{shadish2002experimental} of our evidences and claims.
\textbf{Construct validity} threats consist of concepts misrepresenting the object of study, for instance, assuming that accuracy represents the performance of the classifier when the samples are imbalanced (i.e., unequal number of negative and positive instances) or when false negatives and false positives have distinct levels of importance (i.e., costs, risks/safety) for downstream tasks (e.g., planning). We mitigated this validity threat by working only with balanced datasets. Although sound, this assumption might be too strong for real-world settings, where number of instances, relative costs/risks, and error rates are not homogeneous or comparable across classes~\cite{stockl2022stablediffimagenet}.
%or where the number of classes differs as in the outcomes of Imagenet-100 vs. 1k (see Table 1 in~\cite{sariyildiz2023fake}).
%
\textbf{Internal validity} threats relate to assumptions that can be invalidated by alternative explanations for the measured effects or lack thereof, e.g., hidden confounders or selection bias induced by data leakage. We mitigated this by controlling certain parameters (e.g., freezing vs.\ reinitializing layers), taking the last 5 training epochs, and evaluating the top-1/top-5 outcomes. However, as we were unable to control all the parameters, there might be vestiges of confounding, e.g., (1) how the data augmentation might have produced selection biases (certain types of images benefiting more from the augmentations) or (2) the competing effects of oracle prompts and additional information the image embeddings used to simulate them may contain. A promising avenue is to further probe with the simulated prompts, as the positive effects of prompt engineering are corroborated by similar settings like WaffleCLIP~\cite{roth2023waffling}.
\textbf{External validity} threats correspond to configurations (e.g., hyperparameters) that hinder the generalization/reproduction of similar accuracy gains in future experiments (e.g., under different methods/domains). We mitigated this risk by performing various sensitivity analyses, for instance, varying the intake order of real vs.\ synthetic data in pre-training and the number of frozen layers. Albeit extensive, our evaluations were not exhaustive. Further promising work is to study if different architecture types and domains/classes can induce the phenomenon of underspecification~\cite{damour2020underspecification}, where models with equivalent performance on holdout sets still show distinct outcomes after being deployed.
\subsection{Conclusion and Future Work}
%While catastrophic forgetting is mostly harmful, it could also be beneficial in generative models~\cite{zhang2023forget,heng2024selective}, for instance, to correct previous overfitting, particularly when learned noise features prevent acquiring new concepts or, worse, posing a threat of leaking private/copyrighted data. 
%Mitigating or inducing forgetting in fine-tuned diffusion models is a future topic of research. We are particularly interested in studying fine-tuning and continual learning of diffusion models in the context of autonomous-driving scenarios, where \textit{CF} emerges both in class-incremental learning (distinct sequence classes, e.g., vehicle types) and domain-incremental learning (for the same class but distinct contexts, e.g., day-time, weather conditions)~\cite{witte2023severity}.
% 
We investigated the accuracy gap between models trained on synthetic data vs.\ real data. Looking into the role of data normalization, batch normalization, and data augmentations for synthetic vs.\ real data, we identified improvements that explain part of the observed accuracy gap.  Our results indicate that the final layers of the student model play a significant role in bridging this gap. When pre-training all but the last two layers using synthetic data, we observed only a minor drop in accuracy. Furthermore, we demonstrated that pre-training the majority of the layers with synthetic data and fine-tuning the remaining layers with a fraction of real data resulted in improved performance compared to models trained merely on a subset of real data. These results suggest leveraging synthetic data to mitigate a lack of labeled real training data.
Our findings point to promising new avenues for future research. Firstly, a deeper exploration of the specific features learned by the final layers could shed light on their contribution to the observed accuracy gaps. Additionally, investigating various ratios of synthetic and real data during fine-tuning may reveal better trade-offs, potentially incorporating active learning to select the most informative labeled data. Lastly, an investigation into how different generative foundation models impact student model performance would provide a comprehensive understanding of the potential of synthetic data in transfer learning.
% 
%\begin{credits}
\subsubsection{\ackname} This work was supported in part by Bosch Center for Artificial Intelligence and the Hasso-Plattner-Institute.
We thank Claudia Blaiotta, Bastian Bischoff, Niclas Popp, Anna Khoreva, and Jan H. Metzen for their discussions and feedback on the manuscript. 
The Bosch Group is carbon neutral. Administration, manufacturing and research activities do no longer leave a carbon footprint. This also includes GPU clusters on which most experiments have been performed.
\subsubsection{\discintname} The authors have no competing interests to declare that are relevant to the content of this article.
%It is now necessary to declare any competing interests or to specifically state that the authors have no competing interests. Please place the statement with a bold run-in heading in small font size beneath the
%(optional) %acknowledgments\footnote{If EquinOCS, our proceedings submission system, is used, then the disclaimer can be provided directly in the system.}, for example: The authors have no competing interests to declare that arerelevant to the content of this article. Or: Author A has received research grants from Company W. Author B has received a speaker honorarium from Company X and owns stock in Company Y. Author C is a member of committee Z.
%\end{credits}
\bibliographystyle{splncs04}
\bibliography{references}

\begin{thebibliography}{10}
\providecommand{\url}[1]{\texttt{#1}}
\providecommand{\urlprefix}{URL }
\providecommand{\doi}[1]{https://doi.org/#1}

\bibitem{pytorchExampleAugs}
AI, M.: Pytorch examples. http://github.com/pytorch/examples (2013)

\bibitem{azizi_synthetic_2023}
Azizi, S., Kornblith, S., Saharia, C., Norouzi, M., Fleet, D.J.: Synthetic data from diffusion models improves imagenet classification. arXiv:2304.08466  (2023)

\bibitem{bhardwaj2019edgeal}
Bhardwaj, K., Suda, N., Marculescu, R.: Edgeal: A vision for deep learning in the iot era. IEEE Design \& Test  \textbf{38}(4),  37--43 (2019)

\bibitem{binici2022robust}
Binici, K., Aggarwal, S., Pham, N.T., Leman, K., Mitra, T.: Robust and resource-efficient data-free knowledge distillation by generative pseudo replay. In: AAAI Conf. on Artificial Intelligence. vol.~36, pp. 6089--6096 (2022)

\bibitem{binici2022preventing}
Binici, K., Pham, N.T., Mitra, T., Leman, K.: Preventing catastrophic forgetting and distribution mismatch in knowledge distillation via synthetic data. In: CVF Conf. on Applications of Computer Vision. pp. 663--671 (2022)

\bibitem{bommasani2021opportunitiesfoundation}
Bommasani, R., Hudson, D.A., Adeli, E., Altman, R., Arora, S., von Arx, S., Bernstein, M.S., Bohg, J., Bosselut, A., Brunskill, E., et~al.: On the opportunities and risks of foundation models. arXiv:2108.07258  (2021)

\bibitem{caron_emerging_2021}
Caron, M., Touvron, H., Misra, I., Jégou, H., Mairal, J., Bojanowski, P., Joulin, A.: Emerging {Properties} in {Self}-{Supervised} {Vision} {Transformers}. In: ICCV Intl. Conf. on Computer Vision (2021)

\bibitem{chang2019disjoint}
Chang, X., Yang, Y., Xiang, T., Hospedales, T.M.: Disjoint label space transfer learning with common factorised space. In: AAAI Conf. on Artificial Intelligence. vol.~33, pp. 3288--3295 (2019)

\bibitem{changpinyo_conceptual_2021}
Changpinyo, S., Sharma, P., Ding, N., Soricut, R.: Conceptual 12m: {Pushing} web-scale image-text pre-training to recognize long-tail visual concepts. In: CVPR Conf. on Computer Vision and Pattern Recognition. pp. 3558--3568 (2021)

\bibitem{chen2019data}
Chen, H., Wang, Y., Xu, C., Yang, Z., Liu, C., Shi, B., Xu, C., Xu, C., Tian, Q.: Data-free learning of student networks. In: CVF Intl. Conf. on Computer Vision. pp. 3514--3522 (2019)

\bibitem{damour2020underspecification}
D'Amour, A., Heller, K., Moldovan, D., Adlam, B., Alipanahi, B., Beutel, A., Chen, C., Deaton, J., Eisenstein, J., Hoffman, M.D., et~al.: Underspecification presents challenges for credibility in modern machine learning. Journal of Machine Learning Research  \textbf{23}(226),  1--61 (2022)

\bibitem{deng2009imagenet}
Deng, J., Dong, W., Socher, R., Li, L.J., Li, K., Fei-Fei, L.: Imagenet a large-scale hierarchical image database. In: CVPR Conf. on computer vision and pattern recognition. pp. 248--255 (2009)

\bibitem{desai_redcaps_2021}
Desai, K., Kaul, G., Aysola, Z., Johnson, J.: {RedCaps}: {Web}-curated image-text data created by the people, for the people. arXiv:2111.11431  (2021)

\bibitem{do2022momentum}
Do, K., Le, T.H., Nguyen, D., Nguyen, D., Harikumar, H., Tran, T., Rana, S., Venkatesh, S.: Momentum adversarial distillation: Handling large distribution shifts in data-free knowledge distillation. Advances in Neural Information Processing Systems  \textbf{35},  10055--10067 (2022)

\bibitem{dosovitskiy2020image}
Dosovitskiy, A., Beyer, L., Kolesnikov, A., Weissenborn, D., Zhai, X., Unterthiner, T., Dehghani, M., Minderer, M., Heigold, G., Gelly, S., et~al.: An image is worth 16x16 words: Transformers for image recognition at scale. arXiv:2010.11929  (2020)

\bibitem{geirhos2020shortcut}
Geirhos, R., Jacobsen, J.H., Michaelis, C., Zemel, R., Brendel, W., Bethge, M., Wichmann, F.A.: Shortcut learning in deep neural networks. Nature Machine Intelligence  \textbf{2}(11),  665--673 (2020)

\bibitem{geirhos2018imagenet}
Geirhos, R., Rubisch, P., Michaelis, C., Bethge, M., Wichmann, F.A., Brendel, W.: Imagenet-trained cnns are biased towards texture; increasing shape bias improves accuracy and robustness. arXiv:1811.12231  (2018)

\bibitem{goodfellow2020gans}
Goodfellow, I., Pouget-Abadie, J., Mirza, M., Xu, B., Warde-Farley, D., Ozair, S., Courville, A., Bengio, Y.: Generative adversarial networks. Communications of the ACM  \textbf{63}(11),  139--144 (2020), publisher: ACM New York, NY, USA

\bibitem{gou2021distillingsurvey}
Gou, J., Yu, B., Maybank, S.J., Tao, D.: Knowledge distillation: A survey. Intl. Journal of Computer Vision  \textbf{129}(6),  1789--1819 (2021)

\bibitem{hammoud2024synthclip}
Hammoud, H.A.A.K., Itani, H., Pizzati, F., Torr, P., Bibi, A., Ghanem, B.: Synthclip: Are we ready for a fully synthetic clip training? arXiv:2402.01832  (2024)

\bibitem{he2016deep}
He, K., Zhang, X., Ren, S., Sun, J.: Deep residual learning for image recognition. In: CVPR Conf. on Computer Vision and Pattern Recognition. pp. 770--778 (2016)

\bibitem{hendrycks_augmix_2020}
Hendrycks, D., Mu, N., Cubuk, E.D., Zoph, B., Gilmer, J., Lakshminarayanan, B.: {AugMix}: {A} {Simple} {Data} {Processing} {Method} to {Improve} {Robustness} and {Uncertainty}. ICLR Intl. Conf. on Learning Representations  (2020)

\bibitem{hendrycks_pixmix_2022}
Hendrycks, D., Zou, A., Mazeika, M., Tang, L., Li, B., Song, D., Steinhardt, J.: {PixMix}: {Dreamlike} {Pictures} {Comprehensively} {Improve} {Safety} {Measures}. In: CVPR Conf. on Computer Vision and Pattern Recognition (2022)

\bibitem{hinton2015distilling}
Hinton, G., Vinyals, O., Dean, J., et~al.: Distilling the knowledge in a neural network. arXiv:1503.02531  (2015)

\bibitem{ho2022stablediffguidance}
Ho, J., Salimans, T.: Classifier-free diffusion guidance. arXiv:2207.12598  (2022)

\bibitem{lecun2015deep}
LeCun, Y., Bengio, Y., Hinton, G.: Deep learning. Nature  \textbf{521}(7553) (2015)

\bibitem{liang2018generative}
Liang, K.J., Li, C., Wang, G., Carin, L.: Generative adversarial network training is a continual learning problem. arXiv:1811.11083  (2018)

\bibitem{lopes2017datafree}
Lopes, R.G., Fenu, S., Starner, T.: Data-free knowledge distillation for deep neural networks. arXiv:1710.07535  (2017)

\bibitem{mccloskey1989catastrophic}
McCloskey, M., Cohen, N.J.: Catastrophic interference in connectionist networks: The sequential learning problem. In: Psychology of learning and motivation, vol.~24, pp. 109--165. Elsevier (1989)

\bibitem{miller1995wordnet}
Miller, G.A.: Wordnet: a lexical database for english. Communications of the ACM  \textbf{38}(11),  39--41 (1995)

\bibitem{miotto2018deeplhealthcarechallenges}
Miotto, R., Wang, F., Wang, S., Jiang, X., Dudley, J.T.: Deep learning for healthcare: review, opportunities and challenges. Briefings in bioinformatics  \textbf{19}(6),  1236--1246 (2018)

\bibitem{radford2021learning}
Radford, A., Kim, J.W., Hallacy, C., Ramesh, A., Goh, G., Agarwal, S., Sastry, G., Askell, A., Mishkin, P., Clark, J., et~al.: Learning transferable visual models from natural language supervision. In: ICML Intl. Conf. on Machine Learning. pp. 8748--8763 (2021)

\bibitem{radford2021clip}
Radford, A., Kim, J.W., Hallacy, C., Ramesh, A., Goh, G., Agarwal, S., Sastry, G., Askell, A., Mishkin, P., Clark, J., et~al.: Learning transferable visual models from natural language supervision. In: ICML Intl. Conf. on Machine Learning. pp. 8748--8763 (2021)

\bibitem{ramesh2022hierarchical}
Ramesh, A., Dhariwal, P., Nichol, A., Chu, C., Chen, M.: Hierarchical text-conditional image generation with clip latents. arXiv:2204.06125  \textbf{1}(2), ~3 (2022)

\bibitem{ravi2016deeplchallengeswearable}
Ravi, D., Wong, C., Lo, B., Yang, G.Z.: A deep learning approach to on-node sensor data analytics for mobile or wearable devices. IEEE Journal of Biomedical and Health Informatics  \textbf{21}(1),  56--64 (2016)

\bibitem{rombach2022high}
Rombach, R., Blattmann, A., Lorenz, D., Esser, P., Ommer, B.: High-resolution image synthesis with latent diffusion models. In: CVPR Conf. on Computer Vision and Pattern Recognition. pp. 10684--10695 (2022)

\bibitem{rombach2022stablediff}
Rombach, R., Blattmann, A., Lorenz, D., Esser, P., Ommer, B.: High-resolution image synthesis with latent diffusion models. In: CVPR Conf. on Computer Vision and Pattern Recognition. pp. 10684--10695 (2022)

\bibitem{roth2023waffling}
Roth, K., Kim, J.M., Koepke, A., Vinyals, O., Schmid, C., Akata, Z.: Waffling around for performance: Visual classification with random words and broad concepts. In: CVF Intl. Conf. on Computer Vision. pp. 15746--15757 (2023)

\bibitem{ruder2016overview}
Ruder, S.: An overview of gradient descent optimization algorithms. arXiv:1609.04747  (2016)

\bibitem{saharia2022imagen}
Saharia, C., Chan, W., Saxena, S., Li, L., Whang, J., Denton, E., Ghasemipour, S.K.S., Ayan, B.K., Mahdavi, S.S., Lopes, R.G., et~al.: Photorealistic text-to-image diffusion models with deep language understanding. arXiv:2205.11487  (2022)

\bibitem{sariyildiz2023fake}
Sar{\i}y{\i}ld{\i}z, M.B., Alahari, K., Larlus, D., Kalantidis, Y.: Fake it till you make it: Learning transferable representations from synthetic imagenet clones. In: CVPR Conf. on Computer Vision and Pattern Recognition. pp. 8011--8021 (2023)

\bibitem{sariyildiz2022no}
Sariyildiz, M.B., Kalantidis, Y., Alahari, K., Larlus, D.: No reason for no supervision: Improved generalization in supervised models. arXiv preprint arXiv:2206.15369  (2022)

\bibitem{shadish2002experimental}
Shadish, W.R., Cook, T.D., Campbell, D.T.: Experimental and quasi-experimental designs for generalized causal inference. Houghton, Mifflin and Company (2002)

\bibitem{sharma_conceptual_2018}
Sharma, P., Ding, N., Goodman, S., Soricut, R.: Conceptual captions: {A} cleaned, hypernymed, image alt-text dataset for automatic image captioning. In: ACL Association for Computational Linguistics. pp. 2556--2565 (2018)

\bibitem{stockl2022stablediffimagenet}
St{\"o}ckl, A.: Evaluating a synthetic image dataset generated with stable diffusion. arXiv:2211.01777  (2022)

\bibitem{thanh2020catastrophic}
Thanh-Tung, H., Tran, T.: Catastrophic forgetting and mode collapse in gans. In: IJCNN Intl. Joint Conf. on Neural Networks. pp. 1--10. IEEE (2020)

\bibitem{tian2024stablerep}
Tian, Y., Fan, L., Isola, P., Chang, H., Krishnan, D.: Stablerep: Synthetic images from text-to-image models make strong visual representation learners. Advances in Neural Information Processing Systems  \textbf{36} (2024)

\bibitem{tian2020contrastive}
Tian, Y., Krishnan, D., Isola, P.: Contrastive multiview coding. In: ECCV European Computer Vision Conf. pp. 776--794. Springer (2020)

\bibitem{wang2024comprehensive}
Wang, L., Zhang, X., Su, H., Zhu, J.: A comprehensive survey of continual learning: Theory, method and application. IEEE Transactions on Pattern Analysis and Machine Intelligence  (2024)

\bibitem{wang2023efficient}
Wang, X., Lin, B., Liu, D., Xu, C.: Efficient transfer learning in diffusion models via adversarial noise (2023)

\bibitem{wang2023comprehensive}
Wang, Z., Yang, E., Shen, L., Huang, H.: A comprehensive survey of forgetting in deep learning beyond continual learning. arXiv:2307.09218  (2023)

\bibitem{yamaguchi2022transfer}
Yamaguchi, S., Kanai, S., Kumagai, A., Chijiwa, D., Kashima, H.: Transfer learning with pre-trained conditional generative models. arXiv:2204.12833  (2022)

\bibitem{zhang2019empiricaldeeplchallenges}
Zhang, T., Gao, C., Ma, L., Lyu, M., Kim, M.: An empirical study of common challenges in developing deep learning applications. In: ISSRE Intl. Symp. on Software Reliability Engineering. pp. 104--115 (2019)

\bibitem{zoph2020rethinking}
Zoph, B., Ghiasi, G., Lin, T.Y., Cui, Y., Liu, H., Cubuk, E.D., Le, Q.: Rethinking pre-training and self-training. Advances in neural information processing systems  \textbf{33},  3833--3845 (2020)

\end{thebibliography}
\end{document}

\maketitle
\begin{table}[htbp]
\caption{\textbf{Real Data Transfer Learning.} This table provides the concrete results for Figure 2a. Using the first N layers of the model pre-trained on real ImageNet-100 data with frozen weights, the remaining layers were initialized randomly and trained on synthetic ImageNet-100 data. We report the mean and standard deviation on real ImageNet-100 validation data of the last 5 epochs for top-1 and top-5 accuracy.}
\label{tab:abl_6_real_data_transfer_learning}
\begin{tabular}{ l C{2.5cm} C{2.5cm} C{2.5cm} }
\toprule
Frozen Layers & Top-1 Acc. (\%) & Top-5 Acc.  (\%) & Train Loss \\
\midrule
None - Real Data Only (R) & 87.8 ±0.1 & 97.1 ±0.1 & 0.742 ±0.002 \\
\midrule
None - Synth. Data Only (S) & 64.8 ±0.1 & 87.6 ±0.1 & 0.696 ±0.003 \\

+ L. 0-1\hspace{0.16cm} from (R) & 65.4 ±0.3 & 87.4 ±0.2 & 0.713 ±0.003 \\
+ L. 0-2\hspace{0.16cm} from (R) & 65.0 ±0.1 & 87.4 ±0.1 & 0.723 ±0.003 \\
+ L. 0-3\hspace{0.16cm} from (R) & 64.6 ±0.1 & 87.2 ±0.1 & 0.724 ±0.002 \\
+ L. 0-4\hspace{0.16cm} from (R) & 64.9 ±0.2 & 87.8 ±0.2 & 0.721 ±0.004 \\
+ L. 0-5\hspace{0.16cm} from (R) & 64.7 ±0.2 & 86.4 ±0.1 & 0.727 ±0.004 \\
+ L. 0-6\hspace{0.16cm} from (R) & 64.7 ±0.2 & 86.6 ±0.2 & 0.732 ±0.002 \\
+ L. 0-7\hspace{0.16cm} from (R) & 64.8 ±0.1 & 87.6 ±0.2 & 0.738 ±0.003 \\
+ L. 0-8\hspace{0.16cm} from (R) & 64.3 ±0.2 & 86.6 ±0.2 & 0.740 ±0.003 \\
+ L. 0-9\hspace{0.16cm} from (R) & 65.3 ±0.1 & 86.7 ±0.2 & 0.746 ±0.004 \\
+ L. 0-10 from (R) & 65.7 ±0.2 & 88.3 ±0.2 & 0.748 ±0.002 \\
+ L. 0-11 from (R) & 64.6 ±0.3 & 87.0 ±0.2 & 0.698 ±0.004 \\
+ L. 0-12 from (R) & 65.8 ±0.4 & 87.1 ±0.1 & 0.703 ±0.002 \\
+ L. 0-13 from (R) & 66.3 ±0.4 & 87.6 ±0.2 & 0.712 ±0.001 \\
+ L. 0-14 from (R) & 66.8 ±0.4 & 87.7 ±0.1 & 0.727 ±0.002 \\
+ L. 0-15 from (R) & {68.6 ±0.3} & 89.1 ±0.2 & 0.889 ±0.002 \\
+ L. 0-16 from (R) & {71.6 ±0.2} & 90.5 ±0.2 & 1.069 ±0.002 \\
+ L. 0-17 from (R) & {85.2 ±0.3} & 95.6 ±0.2 & 1.925 ±0.003 \\
\bottomrule
\end{tabular}
\end{table} 

\begin{table}
\caption{\textbf{Synthetic Data Transfer Learning.} This table provides the concrete results for Figure 2b. Using the first N layers of the model pre-trained on synthetic ImageNet-100 data with frozen weights, the remaining layers were initialized randomly and trained on real ImageNet-100 data. We report the mean and standard deviation on real ImageNet-100 validation data of the last 5 epochs for top-1 and top-5 accuracy.}
\label{tab:abl_9_synthetic_data_transfer_learning}
\begin{tabular}{ l C{2.5cm} C{2.5cm} C{3cm} }
\toprule
Frozen Layers & Top-1 Acc. (\%) & Top-5 Acc.  (\%) & Train Loss \\
\midrule
None - Synth. Data Only (S) & 64.8 ±0.1 & 87.6 ±0.1 & 0.696 ±0.003 \\
\midrule
None - Real Data Only (R) & 87.8 ±0.1 & 97.1 ±0.1 & 0.742 ±0.002 \\
+ L. 0-1\hspace{0.16cm} from (S) & 88.5 ±0.2 & 97.4 ±0.1 & 0.713 ±0.003 \\
+ L. 0-2\hspace{0.16cm} from (S) & 88.8 ±0.1 & 98.0 ±0.1 & 0.747 ±0.003 \\
+ L. 0-3\hspace{0.16cm} from (S) & 88.7 ±0.1 & 97.4 ±0.1 & 0.753 ±0.003 \\
+ L. 0-4\hspace{0.16cm} from (S) & 88.2 ±0.1 & 97.5 ±0.1 & 0.752 ±0.002 \\
+ L. 0-5\hspace{0.16cm} from (S) & 88.4 ±0.2 & 97.7 ±0.1 & 0.762 ±0.004 \\
+ L. 0-6\hspace{0.16cm} from (S) & 88.2 ±0.1 & 97.6 ±0.1 & 0.762 ±0.004 \\
+ L. 0-7\hspace{0.16cm} from (S) & 87.6 ±0.1 & 97.5 ±0.1 & 0.763 ±0.003 \\
+ L. 0-8\hspace{0.16cm} from (S) & 87.7 ±0.1 & 97.8 ±0.1 & 0.769 ±0.004 \\
+ L. 0-9\hspace{0.16cm} from (S) & 87.4 ±0.1 & 97.3 ±0.1 & 0.786 ±0.001 \\
+ L. 0-10 from (S) & 88.2 ±0.1 & 97.4 ±0.1 & 0.787 ±0.001 \\
+ L. 0-11 from (S) & 87.5 ±0.1 & 97.6 ±0.1 & 0.797 ±0.003 \\
+ L. 0-12 from (S) & 87.2 ±0.1 & 97.0 ±0.1 & 0.808 ±0.003 \\
+ L. 0-13 from (S) & 87.1 ±0.2 & 97.1 ±0.1 & 0.824 ±0.003 \\
+ L. 0-14 from (S) & 86.8 ±0.1 & 96.7 ±0.1 & 0.822 ±0.002 \\
+ L. 0-15 from (S) & {85.9 ±0.2} & 96.8 ±0.1 & 1.010 ±0.002 \\
+ L. 0-16 from (S) & {84.3 ±0.2} & 96.0 ±0.1 & 1.277 ±0.002 \\
+ L. 0-17 from (S) & {73.0 ±0.2} & 92.1 ±0.1 & 2.104 ±0.004 \\
\bottomrule
\end{tabular}
\end{table}

\begin{table}
\caption{\textbf{Data Reduction Experiments.} This table provides the concrete data points for Figure 4. The models in  \Cref{tab:abl_15_data_reduction_synth} (except for the baselines) were trained with the first 16 layers of the model pre-trained on synthetic ImageNet-100 data with frozen weights and the remaining layers being initialized randomly and trained on real ImageNet-100 data. The real training data used to train the last two layers was reduced to the fraction specified in the table using random sampling. In  \Cref{tab:abl_15_data_reduction_real} we show the results of the same data reduction applied to models that are trained on real data only. We report the mean and standard deviation on real ImageNet-100 validation data of the last 5 epochs for top-1 and top-5 accuracy.}
\label{tab:abl_15_data_reduction}
\begin{subtable}{\textwidth}
    \caption{Synthetic Data Transfer Learning}
    \label{tab:abl_15_data_reduction_synth}
    \begin{tabular}{ l C{2.5cm} C{2.5cm} C{3cm} }
    \toprule
    Experiment & Top-1 Acc. (\%) & Top-5 Acc.  (\%) & Train Loss \\
    \midrule
    Real Data Only (R) & 87.8 ±0.1 & 97.1 ±0.1 & 0.742 ±0.002 \\
    \midrule
    Synth. Data Only (S) & 64.8 ±0.1 & 87.6 ±0.1 & 0.696 ±0.003 \\
    + 1/1\hphantom{c}\hphantom{c} of real data used & {84.3 ±0.2} & 96.0 ±0.1 & 1.277 ±0.002 \\
    + 1/2\hphantom{c}\hphantom{c} of real data used & {83.0 ±0.1} & 96.2 ±0.1 & 1.313 ±0.004 \\
    + 1/4\hphantom{c}\hphantom{c} of real data used & {81.9 ±0.2} & 95.2 ±0.1 & 1.398 ±0.005 \\
    + 1/8\hphantom{c}\hphantom{c} of real data used & {80.1 ±0.2} & 94.7 ±0.1 & 1.477 ±0.006 \\
    + 1/16\hphantom{c} of real data used & {77.3 ±0.3} & 93.8 ±0.1 & 1.533 ±0.007 \\
    + 1/32\hphantom{c} of real data used & {74.9 ±0.3} & 93.0 ±0.1 & 1.558 ±0.006 \\
    + 1/64\hphantom{c} of real data used & {72.5 ±0.2} & 91.6 ±0.1 & 1.616 ±0.021 \\
    + 1/128 of real data used & {68.5 ±0.2} & 90.8 ±0.2 & 1.516 ±0.039 \\
    \bottomrule
    \end{tabular}
    \end{subtable}
    \begin{subtable}{\textwidth}
    \caption{Real Data Only}%. (R) indicate the use of real data for the baseline.}
    \label{tab:abl_15_data_reduction_real}
    \begin{tabular}{ l C{2.5cm} C{2.5cm} C{3cm} }
    \toprule
    Experiment & Top-1 Acc. (\%) & Top-5 Acc.  (\%) & Train Loss \\
    \midrule
    1/1\hphantom{c}\hphantom{c} of real data used & 87.8 ±0.1 & 97.1 ±0.1 & 0.742 ±0.002 \\
    1/2\hphantom{c}\hphantom{c} of real data used & 84.6 ±0.1 & 99.0 ±0.1 & 0.822 ±0.002 \\
    1/4\hphantom{c}\hphantom{c} of real data used & 78.7 ±0.1 & 94.1 ±0.1 & 1.124 ±0.002 \\
    1/8\hphantom{c}\hphantom{c} of real data used & 67.7 ±0.1 & 89.0 ±0.2 & 1.529 ±0.002 \\
    1/16\hphantom{c} of real data used & 49.5 ±0.2 & 78.2 ±0.2 & 2.179 ±0.004 \\
    1/32\hphantom{c} of real data used & 32.3 ±0.1 & 61.9 ±0.1 & 2.808 ±0.005 \\
    1/64\hphantom{c} of real data used & 18.4 ±0.1 & 43.8 ±0.2 & 3.254 ±0.009 \\
    1/128 of real data used & 10.7 ±0.1 & 28.6 ±0.2 & 3.454 ±0.005 \\
    \bottomrule
    \end{tabular}
    \end{subtable}
    % \caption{The models (except for the baselines \todo{True? Where is synthetic data in the baselines?}) were fully trained on a randomly sampled subset of real ImageNet-100 data. The table shows the mean value of the last 5 epochs of training \todo{is training accuracy shown? or validation accuracy?} for the top-1 and top-5 accuracy on real data as well as the training loss. \todo{table is unclear. Is 1/1 data added (+) to Baseline (R)? The + sign is unclear.}}
    \end{table}